\documentclass[final]{anthology-ch} 

\usepackage{booktabs}
\usepackage{graphicx}
\usepackage{CJKutf8}
\usepackage{float}
\usepackage{tabularx}

\title{Ground Truth Generation for Multilingual Historical NLP using LLMs}

\author[1]{Clovis Gladstone}[
  orcid=0009-0005-0142-2824
]

\author[2]{Zhao Fang}[
  orcid=0009-0004-8546-8952
]

\author[3]{Spencer Dean Stewart}[
  orcid=0000-0002-7497-3046
]

\affiliation{1}{ARTFL Project, Romance Languages and Literatures, University of Chicago, Chicago, USA}
\affiliation{2}{Department of History, University of Chicago, Chicago, USA}
\affiliation{3}{Libraries and School of Information Studies, Purdue University, West Lafayette, USA}


\pubyear{2025}
\pubvolume{1}
\pagestart{1}
\pageend{1}
\conferencename{Proceedings of Conference XXX}
\conferenceeditors{Editor1 Editor2}
\doi{00000/00000}  

\begin{document}

\maketitle

\begin{abstract}
Historical and low-resource NLP remains challenging due to limited annotated data and domain mismatches with modern, web-sourced corpora. This paper outlines our work in using large language models (LLMs) to create ground-truth annotations for historical French (16th–20th centuries) and Chinese (1900–1950) texts. By leveraging LLM-generated ground truth on a subset of our corpus, we were able to fine-tune spaCy to achieve significant gains on period-specific tests for part-of-speech (POS) annotations, lemmatization, and named entity recognition (NER). Our results underscore the importance of domain-specific models and demonstrate that even relatively limited amounts of synthetic data can improve NLP tools for under-resourced corpora in computational humanities research.
\end{abstract}

\section{Introduction} 

This paper outlines the work of our lab in using large language models (LLMs) to generate ground-truth datasets to fine-tune Natural Language Processing (NLP) models for historical analysis. Our recent work has focused primarily on analyzing historical French and Chinese texts where we have confronted countless problems with existing NLP tools in working with archaic spellings, obsolete vocabulary, fluid grammar, fragmentary digitization, and for Chinese, traditional characters and rare variants. These historical datasets lack standard training and validation sets, the creation of which are too costly for our digital humanities lab. While our use of historical French and Chinese texts in this paper is the result of our broader research agendas, we nevertheless suggest that bringing these two corpora together in this study allow us to address different issues of NLP for historical research, serving as potential test cases for using LLM-generated annotations to train specialized models in diverse and multilingual settings. The French corpus is comprised of literary texts from the 1500s-1900s where shifting spelling conventions and word forms routinely trip up off-the-shelf taggers and lemmatizers. In Chinese, we focus on texts from 1900 to 1950, a shorter but pivotal time frame when written language was shifting from classical to contemporary writing conventions, which makes part-of-speech (POS) and named entity recognition (NER) tagging especially difficult. These two examples therefore represent how leveraging LLMs for ground truth creation can address the critical need to generate suitable training data for specialized historical contexts where it is otherwise absent. 

\section{Related Work}

This research builds on existing scholarship addressing the longstanding challenge of applying NLP tools in low-resource and historical language settings. \cite{piotrowski2012natural, guldi2023dangerous} Off-the-shelf models, such as those in NLTK or spaCy, are typically trained on contemporary, web-sourced texts. Consequently, their precision lags significantly when applied to historical materials, which often feature different vocabularies, syntax, and genre conventions. \cite{kapan2022fine, humbel2021named, batjargal2014approach} Traditionally, researchers have addressed this problem through methods like iterative self-training. In this approach, a pre-trained NER system tags unlabeled historical text, and its high-confidence spans are treated as "silver" data to retrain the model, gradually improving its quality with each iteration. \cite{wu2009domain, novotny2023people}

The recent expansion of LLMs offers a powerful alternative for generating the silver- and gold-standard datasets needed for this fine-tuning. The efficacy of LLMs for this purpose is a subject of current debate. While some researchers have noted mixed and inconsistent annotation results, a growing body of work finds that careful prompting, the inclusion of contextual cues, and an iterative evaluation process can lead to high performance on a range of NLP tasks, even outperforming some conventional methods. \cite{hiltmann2025ner4all, wang2023gpt, zhang2023llmaaa, gonzalez2023yes, qin2023chatgpt, bamman2024classification, underwood2025impact} This potential for higher accuracy, however, is offset by significant computational and financial costs. For instance, one of our previous studies found that completing the same tasks took 300 to 2,300 times longer wall-clock time with an LLM than with conventional NLP pipelines. \cite{fang2025comparative} Such time and expense make it impractical for many research labs, including our own, to process entire corpora using LLMs.

Therefore, our work joins a growing body of scholarship arguing that "bigger is not always better." \cite{d2023data, klein2025provocationshumanitiesgenerativeai} We identify a more sustainable path forward in using LLMs to create hyper-specific, smaller models that are fine-tuned to process texts from a specific genre or moment in time. While LLMs may be impractical for full-corpus analysis, we demonstrate their utility in strategically fine-tuning efficient NLP tools like spaCy for targeted, high-performance research in historical and multilingual contexts.

\section{Methodology}

Since the goal of this project was to produce models that could accurately process our corpora, we chose to work with real historical texts rather than generating synthetic "historical" samples given that LLMs struggle to represent the past without inserting anachronisms. \cite{underwood2025languagemodelsrepresentpast} Once we had extracted our random samples of historical passages (typically at the sentence level), we then used LLMs to produce our ground truth dataset for finetuning. The process for both the French and Chinese corpus included the following steps:

\subsection{Corpus Selection}
\begin{itemize}
\item \textbf{French:} For the French corpus, we drew 55,000 sentences from the ARTFL-Frantext database,\footnote{\url{https://artfl-project.uchicago.edu/artfl-frantext}} selecting 11,000 sentences per century from the 16th–20th centuries. These sentences were randomly selected across eras to prevent chronological bias, with an attempt to capture the orthographic and morphological diversity that characterized centuries of linguistic change.\cite{rey2013mille}

\item \textbf{Chinese:} For the Chinese corpus, we randomly selected 10,000 sentences from the Shanghai Library Republican Journal corpus.\footnote{\url{https://textual-optics-lab.uchicago.edu/shanghai-library-republican-journal-corpus}} Sentences were selected across this time period to capture the shifts of Chinese written texts during this transitional period as its forms shifted from classical conventions towards its more contemporary form. \cite{liu1995translingual, stewart2025methodology}
\end{itemize}

\subsection{LLM Annotation and Evaluation:}
Once we selected our corpora, we experimented with different LLMs for generating text annotations in the form of token-level JSON annotations. Similar to previous studies, we found that LLMs produce inconsistent sequence tagging,  \cite{qin2023chatgpt} leading us to instead focus on POS for both French and Chinese, lemmatization for French only, and NER for Chinese only. 

Because gold-standard testing and validation datasets do not yet exist for these time periods, our approach to selecting an ideal LLM for annotation was less systematic and more exploratory, relying on iterative experimentation and qualitative assessments of output quality. This included drawing on our own expertise as specialists in working with these historical French and Chinese texts. We also found that some models, especially commercial LLMs, were better equipped to handle tasks such as Chinese tokenization. We recognize that this counters emerging best practices in the field that favors open-source models to ensure reproducibility and accessibility. \cite{alizadeh2025open, ziems2024can}  Our aim, however, was to prioritize accuracy and feasibility in a domain where reliable benchmarks are still lacking. We see this as an initial step toward developing workflows and evaluation criteria that can eventually be transferred to more open and transparent frameworks as the ecosystem of historical-language resources matures.

For the French corpus, we selected OpenAI’s GPT-4o model. To ensure deterministic outputs suitable for our annotation task, we selected a temperature of 0 for our prompt. The model was tasked with performing lemmatization and part-of-speech (POS) tagging, adhering to Universal Dependencies (UD) guidelines, as well as assigning morphological tags based on the French Treebank tagset. The detailed API prompt guiding the French annotation process is available in Appendix A at the end of the document. Total computing time for the 55,000 passages was around 36 hours.

For Chinese, we selected gemini-2.0-flash. We found that LLM-generated results overall were less consistent for Chinese given the complicated nature of Chinese tokenization, even when providing specific instructions based on the Chinese Treebank (CTB) style guidelines.  \cite{xia2000segmentation, xue2005penn} To minimize inconsistencies, we fed each sentence through the API twice at different temperatures (0.1 and 0.7), only keeping the results when both outputs matched exactly. The detailed API prompt guiding the Chinese annotation process is available in Table B at the end of the document. Total computing time was around 20 hours.

Given the generated nature of our annotation, we conducted a formal evaluation of the labeling produced by the LLMs to assess overall accuracy and identify error patterns in the sample we took from the test set. We randomly selected 100 annotated sentences from the test set (which enabled subsequent evaluation of our custom spaCy models' true error rate), manually verified the annotations, and analyzed the errors. For the French set, we selected 20 sentences from each represented century (16th through 20th centuries), for a total sample of 3,517 tokens. As detailed in Table 1, the overall accuracy was quite high: out of 3,517 annotated tokens, we identified 195 labeling errors, resulting in an accuracy of 96.47\% for POS tagging and 97.98\% for lemmatization in the French set. The breakdown by century reveals minimal variation, with POS accuracy consistently running approximately 1\% lower than lemmatization accuracy. In our Chinese test, we achieved an even lower overall error rate, with only 31 labeling errors across 1807 tokens. Among these, 19 were POS tagging errors and 12 were NER tagging errors, resulting in a POS accuracy of 98.95\% and an NER accuracy of 94.26\% (see Table 2). 

\begin{table}[h]
\centering
\begin{tabular}{lcccc}
\hline
Period & Total Tokens & Total Labeling Errors & POS Acc. (\%) & Lemma Acc. (\%) \\
\hline
1500-1600 & 762 & 30 & 97.90 & 98.16 \\
1600-1700 & 759 & 41 & 96.57 & 98.02 \\
1700-1800 & 631 & 27 & 96.99 & 98.73 \\
1800-1900 & 709 & 54 & 94.50 & 97.88 \\
1900-2000 & 656 & 43 & 96.34 & 97.10 \\
\hline
\textbf{All Centuries} & \textbf{3517} & \textbf{195} & \textbf{96.47} & \textbf{97.98} \\
\hline
\end{tabular}
\caption{French POS and Lemma annotation accuracy across...}
\label{tab:accuracy_by_century}
\end{table}
\begin{table}[h]
    \renewcommand{\arraystretch}{1.2}
    \centering 
    \begin{tabular}{l c c c c} 
        \toprule
        Dataset & Total Tokens & Total Labeling Errors & POS Acc. (\%) & NER Acc. (\%) \\
        \midrule
        Shanghai\_Simp & 1,807 & 31 & 98.95 & 94.26 \\
        \bottomrule
    \end{tabular}
    \caption{Chinese POS and NER annotation accuracy}
    \label{tab:chinese_pos_ner}
\end{table}

In the case of the French, many mistakes are systematic and amenable to automatic post-processing. For example, the French word "pas" is regularly tagged as a particle rather than an adverb. We believe such inconsistencies could be significantly reduced with the use of ensemble methods. One approach would involve generating multiple annotations per token (using either a single LLM with different temperature settings or multiple LLMs) and selecting the highest-confidence label when disagreements occur. In contrast, the issues in Chinese stem mostly from segmentation errors, such as 'Tokyo Institute of Technology', which, when split into three tokens, prevent its recognition as a single named entity. This type of error highlights the need for a more refined, domain-specific tokenization method as a first step towards future improvements.

\subsection{Custom Model Training via spaCy:}

With these synthetic annotations, we converted all labels to the Universal Dependencies (UD) schema, correcting any remaining inconsistencies in tag conventions. To bolster representation of infrequent categories—such as interjections (INTJ) and imperative verb forms—we applied data augmentation by doubling existing examples. The resulting corpus was then shuffled and split into training (80\%), development (10\%), and test (10\%) partitions, stratified by century for the French data and by decade for the Chinese data to preserve temporal balance. Models were trained on a Nvidia RTX 6000 ADA.

We next fine-tuned our models using spaCy’s (version 3.8.4) sequence-labeling framework, with early stopping determined by performance on the development set. For French, we used CamemBERT aV2 (111 million parameters) as the foundational language model for fine-tuning the components of our spaCy pipeline. This is a model which builds upon the DeBERTaV3 architecture and offers significant improvements in performance and tokenization for French compared to the original CamemBERT. \cite{antoun2024camembert} For Chinese, we used zh\_core\_web\_lg (500k unique 300-dimensional word vectors) given its speed, size, and robust POS and NER accuracy. Final evaluation comprised POS-tagging accuracy (for French and Chinese), lemmatization (for French) and NER (for Chinese) against a held-out, manually curated subset, thereby quantifying the effectiveness of LLM-derived synthetic ground truth in bootstrapping robust NLP tools for low-resource historical texts.

\subsection{Validation:}

We validated both the original and fine-tuned models using two complementary tests: first, by evaluating each on our withheld historical test set to measure improvements in processing period-specific language; and second, by assessing performance on a modern reference corpus to gauge general-purpose accuracy, using the UD French-Sequoia\footnote{\url{https://github.com/UniversalDependencies/UD_French-Sequoia}} and UD Chinese-PUD\footnote{\url{https://github.com/UniversalDependencies/UD_Chinese-PUD/tree/master}} treebanks, which are primarily trained on web-sourced text (e.g., Wikipedia, news articles), thereby highlighting the fine-tuned model’s gains on historical text while checking for its robustness on contemporary data.

Chinese proved especially challenging because all downstream tasks depend on accurate segmentation, yet spaCy’s Chinese pipelines use third-party tokenizers (e.g., pkuseg, jieba) tuned for modern simplified text that are less reliable on historical or traditional-character corpora. We finetuned our spaCy model on traditional characters but evaluated it on both traditional and simplified versions of the validation set. Since segmentation errors directly cap POS and NER performance, we also include normalized POS and NER scores, defined as

\[
  \text{Normalized\ Score} = \frac{\text{POS\%\ or\ NER\%}}{\text{Token\ F\%}} \times 100
\]

which adjusts each metric by the underlying tokenization accuracy.

\section{Results}

As seen in Table ~\ref{tab:model-evaluation} summarizing the results for the historical ARTFL–Frantext corpus, the fine-tuned historic model outperformed the off-the-shelf transformer model provided by spaCy, achieving 97.20\% POS accuracy and 96.04\% lemmatization accuracy versus 90.97\% and 87.55\%, respectively when working with historical datasets. Conversely, our contemporary validation dataset (UD French-Sequoia) shows that the off-the-shelf model has an edge on the historical model when working with contemporary data as expected: 98.29\% POS and 94.32\% lemma accuracy compared to 94.10\% and 93.82\%. Moreover, a comparison of the models shows a major performance gap for the off-the-shelf model when shifting from contemporary to historical materials: POS and lemmatization accuracy both drop by around 7\%.  

\begin{table}[h]
\renewcommand{\arraystretch}{1.2}
\begin{tabularx}{\textwidth}{l X X c c} 
\toprule
\hline
\textbf{Dataset} & \textbf{Period} & \textbf{Model} & \textbf{POS} & \textbf{Lemma} \\
\hline
\midrule
ARTFL-Frantext & historical & fr\_dep\_news\_trf & 90.97\% & 87.55\% \\
ARTFL-Frantext & historical & historic & 97.20\% & 96.04\% \\
UD French-Sequoia & modern & fr\_dep\_news\_trf & 98.29\% & 94.32\% \\
UD French-Sequoia & modern & historic & 94.10\% & 93.82\% \\
\bottomrule 
\end{tabularx}%
\caption{French Model evaluation results.}
\label{tab:model-evaluation}
\end{table}

As shown in Table~\ref{tab:zh-model-eval}, fine-tuning yields consistent gains across both historical and contemporary Chinese data. On the traditional‐character Shanghai corpus, the historic model boosts POS accuracy from 67.75\% to 72.33\% (normalized POS from 90.21\% to 96.31\%) and raises NER from 33.44\% to 43.98\% (normalized NER from 44.53\% to 58.56\%). On the simplified‐character Shanghai corpus, POS climbs from 72.77\% to 76.94\% (normalized POS from 88.72\% to 93.81\%) and NER from 41.82\% to 53.60\% (normalized NER from 50.99\% to 65.35\%). Contrary to the French corpus which saw a slip in accuracy for the historical model when working with contemporary data, we find that the historical model outperforms the base model for our validation dataset (UD Chinese-PUD treebank), 77.63\% POS versus 71.70\% (normalized POS 86.32\% vs.\ 79.73\%). 

\begin{table}[htbp]
  \centering
  \renewcommand{\arraystretch}{1.2}
  \resizebox{\columnwidth}{!}{%
    \begin{tabular}{llrrrrr}
      \toprule
      \hline
      \textbf{Dataset} & \textbf{Model} & \textbf{Token F1} & \textbf{POS} & \textbf{POS\_Norm} & \textbf{NER} & \textbf{NER\_Norm} \\
      \hline
      \midrule
      Shanghai\_Trad & zh\_core\_web\_lg               & 75.10\% & 67.75\% & 90.21\% & 33.44\% & 44.53\% \\
      Shanghai\_Trad & historic & 75.10\% & 72.33\% & 96.31\% & 43.98\% & 58.56\% \\
      Shanghai\_Simp & zh\_core\_web\_lg               & 82.02\% & 72.77\% & 88.72\% & 41.82\% & 50.99\% \\
      Shanghai\_Simp & historic & 82.02\% & 76.94\% & 93.81\% & 53.60\% & 65.35\% \\
      UD Chinese-PUD      & zh\_core\_web\_lg               & 89.93\% & 71.70\% & 79.73\% & \textemdash    & \textemdash \\
      UD Chinese-PUD      & historic & 89.93\% & 77.63\% & 86.32\% & \textemdash   & \textemdash \\
      \bottomrule
      \hline
    \end{tabular}%
    }
  \caption{Chinese model evaluation results.}
  \label{tab:zh-model-eval}
\end{table}

\begin{figure}[t!]
  \centering
  \includegraphics[width=\linewidth]{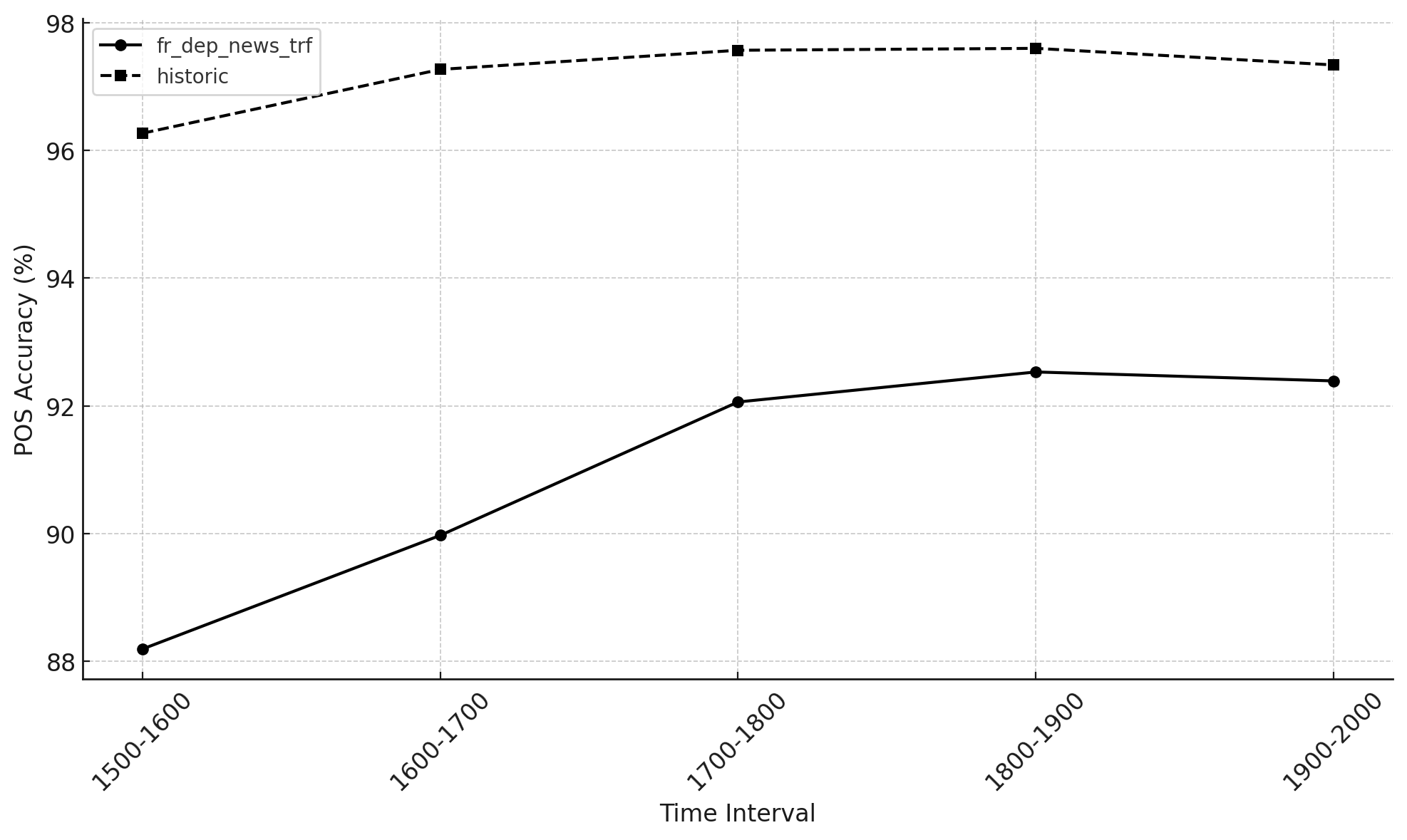}
  \caption{Comparison of POS by model over time.}
  \label{tab:pos_model_over_time}
\end{figure}

The decline in performance that the French corpus saw on the UD French-Sequoia corpus can most likely be explained by several factors: first, the training data is derived from a literary corpus which is quite different from the mostly web/news data used in the Sequoia corpus; second, the coverage area is much larger in the historic model, and in that sense is more generalist than the off-the-shelf spaCy model which is focused on 21st century French. These two observations seem to be verified by a breakdown of performance of both models - historic and contemporary - by century as found in Figure ~\ref{tab:pos_model_over_time}, where we see how the off-the-shelf model struggles with 16th-18th century French, improving gradually as we get closer to contemporary texts, while the historical model achieves impressive consistency over time.

Having identified errors in our LLM-generated training annotations (see section 3.2), we sought to determine whether these imperfections would significantly impact the performance of models trained on this data. To assess this, we evaluated our French corpus given its higher error rate with our custom spaCy model against the standard off-the-shelf spaCy model using our manually validated labels as ground truth. As shown in Table \ref{tab:model_comparison_test_data}, the results demonstrate that the custom historical French model is remarkably robust to the labeling errors in its training data, substantially outperforming the standard modern French model across all time periods, achieving 95.28\% POS accuracy and 96.42\% lemmatization accuracy compared to 92.69\% and 89.28\% respectively—improvements of 2.59 and 7.14 percentage points. The gains are particularly striking in earlier periods: for 16th-century texts, the custom model outperforms the standard model by 6.17 percentage points for POS tagging and 13.91 percentage points for lemmatization.

\begin{table}[H]
\renewcommand{\arraystretch}{1.2}
\begin{tabularx}{\textwidth}{l X X c} 
\toprule
\hline
\textbf{Period} & \textbf{Model} & \textbf{POS Acc (\%)} & \textbf{Lemma Acc (\%)} \\\hline
\midrule
1500-1600 & historic & 95.54 & 93.44 \\
& fr\_dep\_news\_trf & 89.37 & 79.53 \\
\hline
1600-1700 & historic & 96.05 & 96.44 \\
& fr\_dep\_news\_trf & 93.02 & 84.19 \\
\hline
1700-1800 & historic & 95.72 & 97.78 \\
& fr\_dep\_news\_trf & 93.19 & 92.87 \\
\hline
1800-1900 & historic & 95.06 & 98.03 \\
& fr\_dep\_news\_trf & 95.35 & 95.77 \\
\hline
1900-2000 & historic & 93.90 & 96.80 \\
& fr\_dep\_news\_trf & 92.84 & 96.04 \\
\hline
\textbf{All Centuries} & historic & \textbf{95.28} & \textbf{96.42} \\
\textbf{All Centuries} & fr\_dep\_news\_trf & \textbf{92.69} & \textbf{89.28} \\
\hline
\end{tabularx}
\caption{Comparison of modern French and historic French models on validated test data}
\label{tab:model_comparison_test_data}
\end{table}

These results show, albeit with a relatively small sample size, that the error rate in our LLM-generated training annotations does not significantly degrade model performance. The custom model not only learns effectively from imperfect training data but achieves accuracy levels (95-96\%) that closely approach the quality of the training annotations themselves (96-97\%). We believe this robustness to labeling noise validates our approach and suggests that perfect training annotations are not necessary to produce high-quality models for low-resource historical languages.

\section{Discussion}

Our results demonstrate that LLM‐generated synthetic ground truth can effectively bootstrap annotation in lower‐resource historical contexts. Compared to datasets manually annotated by human experts, data generated by LLMs may still be somewhat less accurate. However, the actual error rate proved to be relatively low and is unlikely to materially degrade downstream model performance. This is in part evidenced by the observed improvement to POS results for our historical model when testing the Chinese validation dataset. This suggests that LLMs can serve as a pragmatic source of training data when human‐annotated corpora are scarce or expensive to produce.

Moreover, our results further point to the value and potential of domain‐specific models over a single, all‐purpose pipelines. \cite{gururangan2020don, 10.1145/3458754, beltagy2019scibert, klein2025provocationshumanitiesgenerativeai} Universal models trained on contemporary, web‐sourced text struggle with archaic orthography, obsolete vocabulary, and period‐specific syntax. By contrast, fine‐tuning on even a modest amount of synthetic annotations yields noticeable improvements in POS tagging and NER. In the case of the French model, a clear avenue for future work would be to systematically create and evaluate century-specific models to quantify the potential gains compared the multi-century historic model we built.  We believe that a “many‐models” approach, in which separate pipelines are tailored to particular eras or genres rather than forcing one model to cover every variant of a language, will not only yield higher quality results, but will also demand fewer resources given the smaller amounts of training data needed.  

While this "many-models" approach might not have been possible a few years ago, our study highlights how LLMs make hyper‐specific model development increasingly feasible at scale. With synthetic annotations, researchers are now able to iterate new pipelines for a growing number of historical period or subject domains. Although the optimal amount of synthetic data required will naturally vary based on the specific language, historical period, and NLP tasks (e.g. POS tagging versus NER), our experience suggests that even moderately sized, LLM-generated datasets can provide substantial benefits. Further research in this area could help establish clearer benchmarks for the volume of training data needed for such tasks. There remain certain bottlenecks in place, such as Chinese tokenization, but synthetic annotation might finally be the answer to developing domain-specific tokenizers for languages with notoriously difficult or fluid segmentation practices -- an issue that our team will work on in the future. As LLMs continue to improve, we anticipate that synthetic ground truth will become an indispensable tool for tailoring NLP models to the rich linguistic diversity found in historical corpora. 

\section{Conclusion}

This paper presents a practical method for generating synthetic ground truth via LLMs to bootstrap NLP pipelines in historical contexts. Our experiments on French and Chinese corpora show that fine-tuning on LLM-annotated historical text yields noticeable improvements over off-the-shelf models, even if segmentation accuracy for Chinese remains a critical ceiling for downstream tasks. These findings advocate for a “many-models” strategy, in which separate, era- or genre-tailored pipelines outperform universal solutions. Although our approach depends on the availability of reasonably resource-rich LLMs, it opens new possibilities for computational analysis of lower-resource contexts that lack manual annotations. 

\section*{Limitations}
Although we performed randomized validation checks, which strongly suggest sound quality ground truth data, we acknowledge that our synthetic ground truth data likely still contains subtle or systematic errors. Moreover, our method depends on LLMs extensively trained on modern Chinese and French. While their historical variants remain underrepresented, languages that are not well represented in LLM training data will likely struggle to replicate the results we report.

\section*{Funding Statement}
Computational resources for this project were provided by the ARTFL project at the University of Chicago and a REAL seed grant from the Purdue Libraries and School of Information Studies.

\section*{Disclosure of use of AI tools}
Generative AI was used for assistance in debugging code and in creating some of the figures included in this paper. Otherwise, no AI tools were used outside of the experimental methodology outlined in the paper.

\printbibliography

@article{bamman2024classification,
  title={On Classification with Large Language Models in Cultural Analytics},
  author={Bamman, David and Chang, Kent K and Lucy, Li and Zhou, Naitian},
  journal={arXiv preprint arXiv:2410.12029},
  year={2024}
}

@inproceedings{gonzalez2023yes,
  title={Yes but.. can chatgpt identify entities in historical documents?},
  author={Gonz{\'a}lez-Gallardo, Carlos-Emiliano and Boros, Emanuela and Girdhar, Nancy and Hamdi, Ahmed and Moreno, Jose G and Doucet, Antoine},
  booktitle={2023 ACM/IEEE Joint Conference on Digital Libraries (JCDL)},
  pages={184--189},
  year={2023},
  organization={IEEE}
}

@book{guldi2023dangerous,
  title={The dangerous art of text mining: A methodology for digital history},
  author={Guldi, Jo},
  year={2023},
  publisher={Cambridge University Press}
}

@article{qin2023chatgpt,
  title={Is ChatGPT a general-purpose natural language processing task solver?},
  author={Qin, Chengwei and Zhang, Aston and Zhang, Zhuosheng and Chen, Jiaao and Yasunaga, Michihiro and Yang, Diyi},
  journal={arXiv preprint arXiv:2302.06476},
  year={2023}
}

@book{liu1995translingual,
  title={Translingual practice: Literature, national culture, and translated modernity—China, 1900-1937},
  author={Liu, Lydia He},
  year={1995},
  publisher={Stanford University Press}
}

@article{stewart2025methodology,
  title={A Methodology for Studying Linguistic and Cultural Change in China, 1900-1950},
  author={Stewart, Spencer Dean},
  journal={arXiv preprint arXiv:2502.04286},
  year={2025}
}

@article{wang2023gpt,
  title={Gpt-ner: Named entity recognition via large language models},
  author={Wang, Shuhe and Sun, Xiaofei and Li, Xiaoya and Ouyang, Rongbin and Wu, Fei and Zhang, Tianwei and Li, Jiwei and Wang, Guoyin},
  journal={arXiv preprint arXiv:2304.10428},
  year={2023}
}

@book{piotrowski2012natural,
  title={Natural language processing for historical texts},
  author={Piotrowski, Michael},
  year={2012},
  publisher={Morgan \& Claypool Publishers}
}

@article{fang2025comparative,
  title={A Comparative Analysis of Word Segmentation, Part-of-Speech Tagging, and Named Entity Recognition for Historical Chinese Sources, 1900-1950},
  author={Fang, Zhao and Wu, Liang-Chun and Kong, Xuening and Stewart, Spencer Dean},
  journal={arXiv preprint arXiv:2503.19844},
  year={2025}
}

@article{zhang2023llmaaa,
  title={Llmaaa: Making large language models as active annotators},
  author={Zhang, Ruoyu and Li, Yanzeng and Ma, Yongliang and Zhou, Ming and Zou, Lei},
  journal={arXiv preprint arXiv:2310.19596},
  year={2023}
}

@article{hiltmann2025ner4all,
  title={NER4all or Context is All You Need: Using LLMs for low-effort, high-performance NER on historical texts. A humanities informed approach},
  author={Hiltmann, Torsten and Dr{\"o}ge, Martin and Dresselhaus, Nicole and Grallert, Till and Althage, Melanie and Bayer, Paul and Eckenstaler, Sophie and Mendi, Koray and Schmitz, Jascha Marijn and Schneider, Philipp and others},
  journal={arXiv preprint arXiv:2502.04351},
  year={2025}
}

@article{xia2000segmentation,
  title={The segmentation guidelines for the Penn Chinese Treebank (3.0)},
  author={Xia, Fei},
    journal={University of Pennsylvania Technical Report, IRCS00‐06},
  year={2000}
}

@misc{underwood2025languagemodelsrepresentpast,
      title={Can Language Models Represent the Past without Anachronism?}, 
      author={Ted Underwood and Laura K. Nelson and Matthew Wilkens},
      year={2025},
      eprint={2505.00030},
      archivePrefix={arXiv},
      primaryClass={cs.CL}, 
}

@article{xue2005penn,
  title={The penn chinese treebank: Phrase structure annotation of a large corpus},
  author={Xue, Naiwen and Xia, Fei and Chiou, Fu-Dong and Palmer, Marta},
  journal={Natural language engineering},
  volume={11},
  number={2},
  pages={207--238},
  year={2005},
  publisher={Cambridge University Press}
}

@misc{antoun2024camembert,
  title        = {CamemBERT 2.0: {A} Smarter French Language Model Aged to Perfection},
  author       = {Antoun, Wissam and Kulumba, Francis and Touchent, Rian and de la Clergerie, Éric and Sagot, Benoît and Seddah, Djamé},
  year         = {2024},
  eprint       = {2411.08868},
  archivePrefix = {arXiv},
  primaryClass = {cs.CL},
}

@article{gururangan2020don,
  title={Don't stop pretraining: Adapt language models to domains and tasks},
  author={Gururangan, Suchin and Marasovi{\'c}, Ana and Swayamdipta, Swabha and Lo, Kyle and Beltagy, Iz and Downey, Doug and Smith, Noah A},
  journal={arXiv preprint arXiv:2004.10964},
  year={2020}
}

@article{10.1145/3458754,
author = {Gu, Yu and Tinn, Robert and Cheng, Hao and Lucas, Michael and Usuyama, Naoto and Liu, Xiaodong and Naumann, Tristan and Gao, Jianfeng and Poon, Hoifung},
title = {Domain-Specific Language Model Pretraining for Biomedical Natural Language Processing},
year = {2021},
issue_date = {January 2022},
publisher = {Association for Computing Machinery},
address = {New York, NY, USA},
volume = {3},
number = {1},
doi = {10.1145/3458754},
journal = {ACM Trans. Comput. Healthcare},
month = oct,
articleno = {2},
numpages = {23}
}

@article{beltagy2019scibert,
  title={SciBERT: A pretrained language model for scientific text},
  author={Beltagy, Iz and Lo, Kyle and Cohan, Arman},
  journal={arXiv preprint arXiv:1903.10676},
  year={2019}
}

@article{ziems2024can,
  title={Can large language models transform computational social science?},
  author={Ziems, Caleb and Held, William and Shaikh, Omar and Chen, Jiaao and Zhang, Zhehao and Yang, Diyi},
  journal={Computational Linguistics},
  volume={50},
  number={1},
  pages={237--291},
  year={2024},
  publisher={MIT Press One Broadway, 12th Floor, Cambridge, Massachusetts 02142, USA~…}
}

@misc{klein2025provocationshumanitiesgenerativeai,
      title={Provocations from the Humanities for Generative AI Research}, 
      author={Lauren Klein and Meredith Martin and André Brock and Maria Antoniak and Melanie Walsh and Jessica Marie Johnson and Lauren Tilton and David Mimno},
      year={2025},
      eprint={2502.19190},
      archivePrefix={arXiv},
      primaryClass={cs.CY}, 
}

@article{underwood2025impact,
  title={The impact of language models on the humanities and vice versa},
  author={Underwood, Ted},
  journal={Nature Computational Science},
  pages={1--3},
  year={2025},
  publisher={Nature Publishing Group US New York}
}

@book{d2023data,
  title={Data feminism},
  author={D'ignazio, Catherine and Klein, Lauren F},
  year={2023},
  publisher={MIT press}
}

@inproceedings{wu2009domain,
  title={Domain adaptive bootstrapping for named entity recognition},
  author={Wu, Dan and Lee, Wee Sun and Ye, Nan and Chieu, Hai Leong},
  booktitle={EMNLP'09 Proceedings of the 2009 Conference on Empirical Methods in Natural Language Processing, Volume 3},
  pages={1523--1532},
  year={2009},
  organization={Association for Computing Machinery}
}

@inproceedings{novotny2023people,
  title={People and Places of Historical Europe: Bootstrapping Annotation Pipeline and a New Corpus of Named Entities in Late Medieval Texts},
  author={Novotn{\`y}, V{\'\i}t and Luger, Kristina and {\v{S}}tef{\'a}nik, Michal and Vrabcova, Tereza and Hor{\'a}k, Ale{\v{s}}},
  booktitle={Findings of the Association for Computational Linguistics: ACL 2023},
  pages={14104--14113},
  year={2023}
}

@book{rey2013mille,
  title={Mille ans de langue fran{\c{c}}aise, histoire d’une passion},
  author={Rey, Alain and Duval, Fr{\'e}d{\'e}ric and Siouffi, Gilles},
  year={2013},
  publisher={Perrin}
}

@article{alizadeh2025open,
  title={Open-source LLMs for text annotation: a practical guide for model setting and fine-tuning},
  author={Alizadeh, Meysam and Kubli, Ma{\"e}l and Samei, Zeynab and Dehghani, Shirin and Zahedivafa, Mohammadmasiha and Bermeo, Juan D and Korobeynikova, Maria and Gilardi, Fabrizio},
  journal={Journal of Computational Social Science},
  volume={8},
  number={1},
  pages={17},
  year={2025},
  publisher={Springer}
}

@inproceedings{batjargal2014approach,
  title={An approach to named entity extraction from historical documents in traditional Mongolian script},
  author={Batjargal, Biligsaikhan and Khaltarkhuu, Garmaabazar and Kimura, Fuminori and Maeda, Akira},
  booktitle={IEEE/ACM Joint Conference on Digital Libraries},
  pages={489--490},
  year={2014},
  organization={IEEE}
}

@inproceedings{kapan2022fine,
  title={Fine-tuning NER with spaCy for transliterated entities found in digital collections from the multilingual Persian Gulf},
  author={Kapan, Almazhan and Kirmizialtin, Suphan and Kukreja, Rhythm and Wrisley, David Joseph},
  year={2022},
  organization={CEUR Workshop Proceedings}
}

@article{humbel2021named,
  title={Named-entity recognition for early modern textual documents: a review of capabilities and challenges with strategies for the future},
  author={Humbel, Marco and Nyhan, Julianne and Vlachidis, Andreas and Sloan, Kim and Ortolja-Baird, Alexandra},
  journal={Journal of Documentation},
  volume={77},
  number={6},
  pages={1223--1247},
  year={2021},
  publisher={Emerald Publishing Limited}
}

\appendix

\section{LLM Prompts for French} \label{appdx:first}

\renewcommand{\arraystretch}{1.25} 
\scriptsize

\begingroup
\ttfamily

Parse this French sentence from the 16\textsuperscript{th}--18\textsuperscript{th} centuries into spaCy training format. For each token provide:

\begin{enumerate}
  \item \textbf{Universal Dependencies POS tag (\texttt{pos}):}  
  \texttt{ADJ, ADP, ADV, AUX, CCONJ, DET, INTJ, NOUN, NUM, PART, PRON, PROPN, PUNCT, SCONJ, SYM, VERB, X}

  \item \textbf{French TreeBank tag (\texttt{tag}): Use only the following:}
  \begin{itemize}
    \item \texttt{ADJ} – adjective
    \item \texttt{ADJWH} – interrogative adjective
    \item \texttt{ADV} – adverb
    \item \texttt{ADVWH} – interrogative adverb
    \item \texttt{CC} – coordinating conjunction
    \item \texttt{CLO} – object clitic pronoun
    \item \texttt{CLR} – reflexive clitic pronoun
    \item \texttt{CLS} – subject clitic pronoun
    \item \texttt{CS} – subordinating conjunction
    \item \texttt{DET} – determiner
    \item \texttt{DETWH} – interrogative determiner
    \item \texttt{X} – foreign word
    \item \texttt{I} – interjection
    \item \texttt{NC} – common noun
    \item \texttt{NPP} – proper noun
    \item \texttt{P} – preposition
    \item \texttt{P+D} – preposition+determiner
    \item \texttt{P+PRO} – preposition+pronoun
    \item \texttt{PONCT} – punctuation
    \item \texttt{PREF} – prefix
    \item \texttt{PRO} – pronoun
    \item \texttt{PROREL} – relative pronoun
    \item \texttt{PROWH} – interrogative pronoun
    \item \texttt{V} – verb
    \item \texttt{VIMP} – imperative verb
    \item \texttt{VINF} – infinitive verb
    \item \texttt{VPP} – past participle
    \item \texttt{VPR} – present participle
    \item \texttt{VS} – subjunctive verb
  \end{itemize}

  \item \textbf{Lemma:} Provide in modernized form.

  \item \textbf{Dependency relation (\texttt{dep})}

  \item \textbf{IOB entity tag (\texttt{ent}):} Use \texttt{O} for non-entities.
\end{enumerate}

\textbf{Sentence:} \texttt{\{sentence\}}

\textbf{Required output format:}

\{

  "tokens": [
  
    \{"text": "word", "pos": "NOUN", "tag": "NC", "lemma": "base", "dep": "relation", "ent": "O"\},
    ...
    
  ]
  
\}

Ensure you use exactly these tag sets and maintain historical accuracy in analysis.

\begin{CJK}{UTF8}{bsmi}
\renewcommand{\arraystretch}{1.25} 
\scriptsize

\section{LLM Prompts for Chinese} \label{appdx:first}

\begingroup
\ttfamily

You are a highly precise historical Chinese tokenizer specialized in texts from the 1900–1950 period. Your task is to segment a given sentence according to strict Chinese Treebank (CTB) style guidelines adapted for this historical context, assign accurate Universal Dependencies (UD) POS tags and CTB tags, and perform Named Entity Recognition (NER) using the IOB scheme. **It is crucial to diligently apply NER tagging for all tokens that fit the specified entity types.**\\

\vspace{0.5em}
**Input:**

A single sentence from the 1900–1950 period as plain text.\\

\vspace{0.5em}
**Output:**

Generate **ONLY** a JSON object containing exactly two keys:

1.  `"text"`: The original input sentence string, unchanged.

2.  `"tokens"`: An array of token objects. Each token object MUST have the following keys:

    * `"text"`: The token string.
    
    * `"pos"`: The corresponding Universal Dependencies POS tag. Choose exclusively from: `ADJ`, `ADP`, `ADV`, `AUX`, `CCONJ`, `DET`, `INTJ`, `NOUN`, `NUM`, `PART`, `PRON`, `PROPN`, `PUNCT`, `SCONJ`, `SYM`, `VERB`, `X`.
    
    * `"tag"`: The corresponding Chinese Treebank (CTB) tag. Choose exclusively from: `AD`, `AS`, `BA`, `CC`, `CD`, `CS`, `DEC`, `DEG`, `DER`, `DEV`, `DT`, `ETC`, `FW`, `IJ`, `JJ`, `LB`, `LC`, `M`, `MSP`, `NN`, `NR`, `NT`, `OD`, `ON`, `P`, `PN`, `PU`, `SB`, `SP`, `VA`, `VC`, `VE`, `VV`, `X`.

    * `"ent\_iob\_"`: The IOB tag for NER. Must be one of `"B"` (beginning), `"I"` (inside), or `"O"` (outside). Use `"B"` for the first token of any entity (including single-token entities) and `"I"` for subsequent tokens within the same multi-token entity. Use `"O"` for tokens not part of any named entity.

    * `"ent\_type\_"`: The entity type label. Choose exclusively from: `CARDINAL`, `DATE`, `EVENT`, `FAC`, `GPE`, `LANGUAGE`, `LAW`, `LOC`, `MONEY`, `NORP`, `ORDINAL`, `ORG`, `PERCENT`, `PERSON`, `PRODUCT`, `QUANTITY`, `TIME`, `WORK\_OF\_ART`. Use an empty string `""` if the token is not part of a named entity (`ent\_iob\_` is "O").

\vspace{0.5em}
**Mandatory Tagging Guidelines for Specific Words/Structures:**

* Demonstratives/Proximal Pronouns like "此", "這": Use `pos: PRON`, `tag: PN`. (Do not use DET/DT).

* Third-person/Objective Pronoun "之": Use `pos: PRON`, `tag: PN`.

* Attributive particle "之" (similar to modern 的): Use `pos: PART`, `tag: DEG`. (Do not use DEC).

* Copular verb "爲" (meaning "to be"): Use `pos: VERB`, `tag: VC`. (Distinguish from preposition "爲" `P`).

* Ordinal numbers like "第一": Use `pos: NUM`, `tag: OD`. (Do not use ADJ).

* Localizers like "上", "下", "中": When used post-nominally as locative complements, use `pos: ADP`, `tag: LC`.

* Adjectives like "棘手", "難", "絕妙": Use `pos: ADJ`. For TAG, use `VA` if they can function predicatively or adverbially, `JJ` otherwise. Based on typical CTB, use `VA` for `棘手`, `難`, `絕妙`.

* Particles like "的": Use `DEG` for attributive, `DEC` for relative clause/nominalization marker, `DEV` for V-de constructions. Pay close attention to the historical usage.

* Suffix/Particle "兒": Use `pos: PART`, `tag: ETC`.

* **Religious/National/Political Affiliations (NORP):** Terms referring to specific nationalities (e.g., "美國人" - American), religious affiliations (e.g., "回教" - Islam/Muslim), or political groups (e.g., "國民黨" - Kuomintang) should be identified as `NORP` entities. These terms should generally be tagged as `pos: PROPN`, `tag: NR`. Specifically, for the token "回教" when it refers to the religion or its adherents, the expected output is: `{"text": "回教", "pos": "PROPN", "tag": "NR", "ent\_iob\_": "B", "ent\_type\_": "NORP"}`. This applies even when it modifies another noun (e.g., "回教將軍").

* Titles of Works (often in `「」` or `《》`): Treat the title content as a named entity. Use `pos: PROPN`, `tag: NR` (preferred) or `NN`. Assign `ent\_type\_: WORK\_OF\_ART` for artistic works (books, movies, songs). For famous phrases or political slogans that are not artistic works but might be quoted, the internal tokens should generally be `O` unless they fit another specific entity type from the allowed list. The surrounding quotation marks (`「」`, `《》`, `<>`) themselves should *always* be split off as separate `PUNCT`/`PU` tokens (see Rule 6).

\vspace{0.5em}
**Segmentation Rules (Apply strictly in this priority order):**

1.  **Fixed Multi-char Words \& Set Phrases:**

    * Do NOT split common fixed expressions or grammatical markers prevalent in the period. The provided list `{“是否”, “以後”, “不但”, “其實”, …}` is illustrative and non-exhaustive; apply knowledge of common collocations from the 1900-1950 period. This rule takes precedence over compositional splitting (Rule 4, 5).

2.  **Named Entities \& Acronyms:**

    * Keep recognized geographic names (GPE), personal names (PERSON), organizational names (ORG), specific dates (DATE), times (TIME), etc., as single tokens *if they function as a single unit and are typically written without internal spaces*. Specific date/time formats might be handled differently (see Rule 5). Treat titles of works as potential multi-token entities (`WORK\_OF\_ART`).

3.  **Bound Morphemes (Verbs):**

    * Keep verbs together with their resultative or directional complements (e.g., “看見”, “起來”).

4.  **Compositional Compounds (ADJ+N, N+N):**

    * Split ADJ+N (e.g., “紅” + “圈”, “難” + “問題”) or N+N compounds (e.g., “鈴” + “聲”, “雲” + “半”) *by default if* the meaning is clearly compositional (the first element modifies the second, and potentially other modifiers could be inserted), **unless** the compound is a strongly fixed phrase (Rule 1), a named entity (Rule 2), or a highly lexicalized word where the components don't retain independent meaning (e.g., "政治", "國家" usually stay together). Prioritize splitting when in doubt about lexicalization.

5.  **Other CTB Heuristics:**

    * **Numeral + Measure Word/Noun (CD+M/N):**

        * **ALWAYS split** Numeral (CD) + Measure Word (M) into two separate tokens. Examples: “一” + “群”, “兩” + “條”, “一” + “片”, “一” + “張”, “一” + “則”. The numeral gets `pos: NUM`, `tag: CD`; the measure word gets `pos: NOUN` (or PART depending on context), `tag: M`.

        * Split Numeral (CD) + Noun (N) (e.g., "三" + "人") unless the combination forms a recognized Named Entity (Rule 2, e.g., specific dates like "五月四日" which would be a `DATE`) or a highly conventional fixed unit (e.g., "一體").

    * **Time Expressions:** Tokenize time expressions like "七點鐘" by splitting the numeral from the time unit: "七" (NUM/CD) + "點鐘" (NOUN/NN). Tag the sequence as a `TIME` entity (e.g., `七`/B-TIME, `點鐘`/I-TIME). Simpler forms like "三點" split similarly: "三" (NUM/CD) + "點" (NOUN/NN), potentially tagged as `TIME`.

    * **Reduplication (AA/AABB):** Split into individual components (e.g., "看看" -> "看" + "看") unless the reduplicated form is itself a recognized lexicalized word acting as a single unit (e.g., "隱隱").

    * **Verb-Object (V-O):** **Prioritize splitting** V-O constructions into Verb + Object tokens unless they form a highly lexicalized single concept or idiom (e.g., '革命' might stay together, but split "看" + "書", "寫" + "驢", "見" + "驢").

    * **Verb-Verb (V-V):** Generally split sequences of verbs unless they form a recognized compound verb (e.g., split "愛" + "誦").

6.  **Particles, Measure Words \& Punctuation:**

    * **ALWAYS split** off particles (`DEG`, `DEC`, `DEV`, `AS`, `SP`, `ETC`, etc.), measure words (`M`), localizers (`LC`), and punctuation (`PU`) as separate tokens. This explicitly includes standard punctuation marks (。, ,, ？, !) AND all types of brackets: `「`, `」`, `《`, `》`, `<`, `>`. The content *inside* any brackets should be tokenized according to all other rules (1-5) and NER tagged according to the allowed entity types and specific guidelines.

\vspace{0.5em}
**Final Output Constraint:**

* Ensure the entire output is **only** the JSON object described above. It must start with `{` and end with `}`. Do not include markdown formatting, introductory text, or any explanations outside the JSON structure.

**Sentence to Process:**

\endgroup
\end{CJK}

\end{document}